\documentclass[10pt,twocolumn,letterpaper]{article}

\usepackage{iccv}
\usepackage{times}
\usepackage{epsfig}
\usepackage{graphicx}
\usepackage{amsmath}
\usepackage{amssymb}

\usepackage{url}
\usepackage{soul}
\usepackage{enumerate}
\usepackage{graphicx} 
\usepackage{wrapfig}
\usepackage[inline]{enumitem}
\usepackage{booktabs}
\usepackage{multirow}
\usepackage{pifont}
\newcommand{\cmark}{\ding{51}}%
\newcommand{\xmark}{\ding{55}}%
\usepackage{color}
\usepackage{colortbl}
\definecolor{mygray3}{gray}{.9} 
\definecolor{mygray2}{gray}{.8}
\definecolor{mygray1}{gray}{.7}
\definecolor{myblue}{rgb}{0.61, 0.87, 1.0}
\usepackage{amssymb}
\usepackage{amsfonts,amssymb}

\usepackage[breaklinks=true,bookmarks=false]{hyperref}

\iccvfinalcopy 

\def\iccvPaperID{****} 

\begin{document}
	
	\title{Generative Transition Mechanism to Image-to-Image Translation\\ via Encoded Transformation}

	\author{Yaxin Shi\\
		University of Technology Sydney\\
		{\tt\small Yaxin.Shi@student.uts.edu.au}
		\and
		Xiaowei Zhou\\
		University of Technology Sydney\\
		{\tt\small Xiaowei.Zhou@student.uts.edu.au}
		\and Ping Liu \\
		Institute of High Performance Computing\\
		{\tt\small pino.pingliu@gmail.com}		
		\and Ivor Tsang \\
		University of Technology Sydney\\
		{\tt\small Ivor.Tsang@uts.edu.au}
	}
	
	\maketitle
	\ificcvfinal\thispagestyle{empty}\fi
	
	\begin{abstract}

		{In this paper, we revisit the Image-to-Image (I2I) translation problem with transition consistency, \textit{namely} the consistency defined on the conditional data mapping between each data pairs. Explicitly parameterizing each data mappings with a transition variable $t$, \textit{i.e.,} {\small$x \overset{t(x,y)}{\mapsto}y$}, we discover that existing I2I translation models mainly focus on maintaining {consistency on results}, \textit{e.g.}, image reconstruction or attribute prediction, named result consistency in our paper. This restricts their generalization ability to generate satisfactory results with unseen transitions in the test phase.}
		
		Consequently, we propose to enforce both result consistency and transition consistency for I2I translation, to benefit the problem with a closer consistency between the input and output. To benefit the generalization ability of the translation model, we propose transition encoding to facilitate explicit regularization of these two {kinds} of consistencies on unseen transitions. We further generalize such explicitly regularized consistencies to distribution-level, thus facilitating a generalized overall consistency constraints for I2I translation problems. With the above design, our proposed model, named Transition Encoding GAN (TEGAN), can poss superb generalization ability to generate realistic and semantically consistent translation results with unseen transitions in the test phase. It also provides a unified understanding on the existing GAN-based I2I transition models
		with our explicitly modeling of the data mapping, \textit{i.e.,} transition. Experiments on four different I2I translation tasks demonstrate the efficacy and generality of TEGAN.

		\vspace{-2mm}
	\end{abstract}
	\section{Introduction}
	Image-to-Image (I2I) translation~\cite{DBLP:conf/cvpr/IsolaZZE17} targets to map an image from a source domain $X$ to a target domain $Y$, with these two domains differentiated by different sets of attributes. Under this definition, various computer vision problems can be formulated as I2I translation problems, such as face editing~\cite{laffont2014transient}, style transfer~\cite{DBLP:conf/cvpr/GatysEB16}, image inpainting~\cite{DBLP:conf/cvpr/PathakKDDE16,DBLP:conf/iccv/YangDLYY19}. Recently, following the development of generative adversarial networks (GANs)~\cite{goodfellow2014generative,DBLP:conf/aaai/PanL0YY20}, I2I translation problems have received significant attention. 
	
	In previous I2I translation works, conditional GANs~\cite{goodfellow2014generative} are widely utilized to make translated images with specific attributes as expected. 
	For example, in face editing tasks, RelGAN~\cite{DBLP:conf/iccv/LinWCCL19} utilizes a conditional GAN-based framework to translate an image with an original expression (attributes) to a new image with a different expression. In other words, a successful trained I2I network must make the attributes of translated results consistent with attributes expected. To achieve this goal, previous works either adopt corresponding attribute classifiers to control the attributes of translated results~\cite{DBLP:journals/tip/HeZKSC19,DBLP:conf/cvpr/ChoiCKH0C18,DBLP:conf/nips/ZhuZPDEWS17,DBLP:conf/icml/AlmahairiRSBC18,DBLP:conf/iccv/LinWCCL19,DBLP:conf/cvpr/GongLCG19,hu2020unsupervised}, or design specific loss function,~\textit{e.g.}, cycle-GAN~\cite{DBLP:conf/iccv/ZhuPIE17} to control the consistency between translated results and target domain Y. Those previous works focus on keeping consistency on translated results, which we shortly call~\textit{result consistency}.  
	
\begin{figure}[t]
		\begin{center}
			\vspace{-4mm}
			\includegraphics[width=7.3cm]{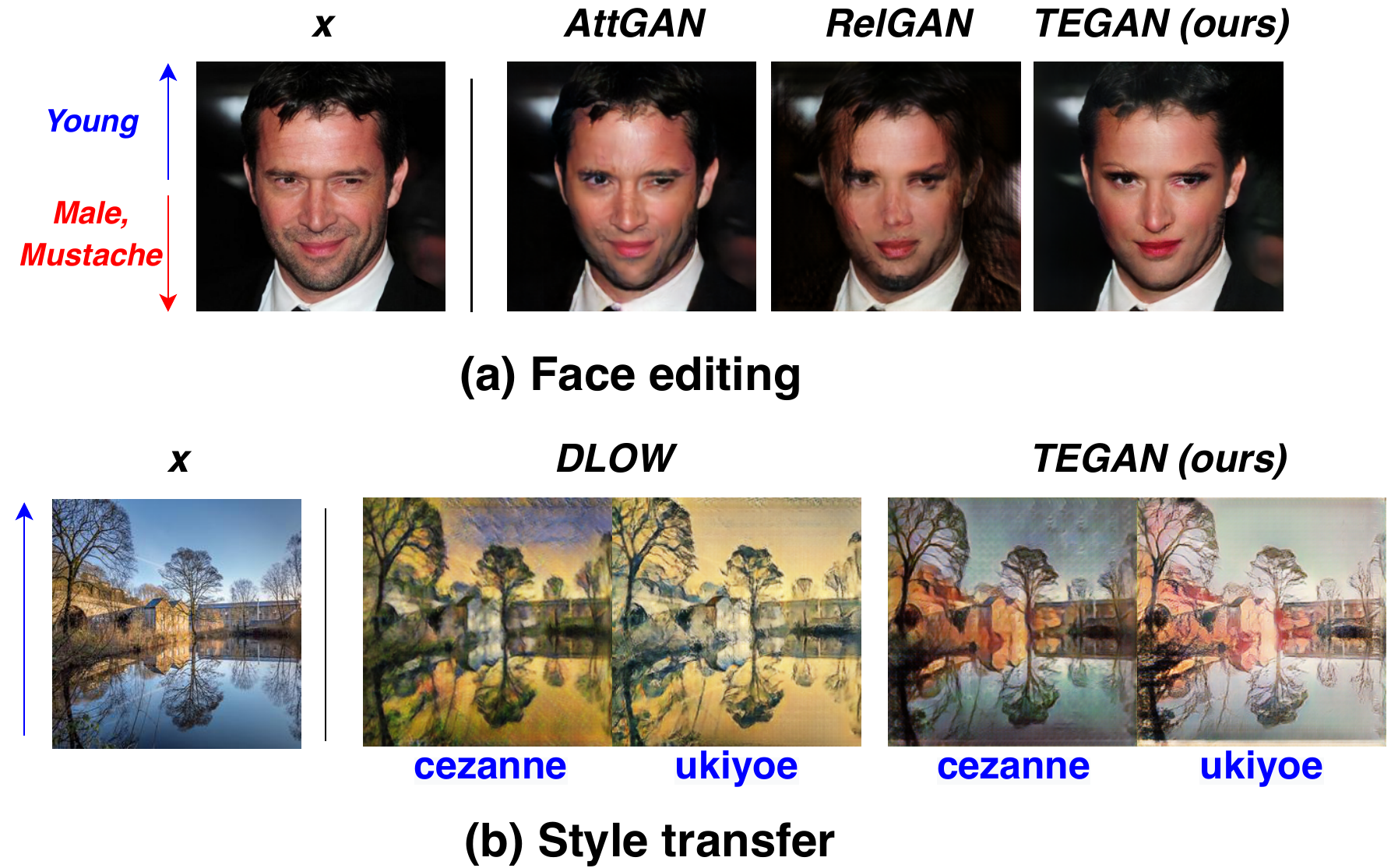}
		\end{center}
		\vspace{-4mm}
		\caption{\small {Example results of our TEGAN. (a) In face editing, our TEGAN outperformed two representative works, \textit{i.e.}, RelGAN, AttGAN. Comparing to RelGAN, our generated results are more realistic while RelGAN has artifacts in its results; comparing to AttGAN, although AttGAN generates results with higher reality than RelGAN, it fails to model the relation between labels and generates unreasonable results, \textit{e.g.}, mustache still exists when we change the gender from male to female.
				(b) In style transfer, our results show distinguishable style for different target domains, \mbox{while the results of DLOW have an apparent defect of style fusion}.
			} 
		}\label{fig:vip_face_editing} 
		\vspace{-6mm}
	\end{figure}

	When looking closer at the I2I translation works focusing on maintaining result consistencies~\cite{DBLP:journals/tip/HeZKSC19,DBLP:conf/cvpr/ChoiCKH0C18,DBLP:conf/nips/ZhuZPDEWS17,DBLP:conf/icml/AlmahairiRSBC18,DBLP:conf/iccv/LinWCCL19,DBLP:conf/cvpr/GongLCG19,DBLP:conf/iccv/ZhuPIE17}, we argue that other than the result consistency in I2I translation, there is another consistency existing while ignored in previous works, which is named as~\textit{transition consistency}. The definition of transition consistency is as follows: if a translated output $\hat{y}$ and $y$ sharing the same attributes (\textit{i.e.} result consistency) are close, they should also share a common conditional transformation. To 
	formulate the~\textit{transition consistency} existing in I2I translation, we explicitly parameterize data transformation between a source data $x$ and a target data $y$ by introducing a \textit{transition} variable $t$, namely {$t\triangleq {t(x,y)}$ }. In this way, an I2I translation problem is reformulated to maintain two consistencies in learning processes,~\textit{i.e.}, result consistency and transition consistency. Considering both result consistency and transition consistency to solve I2I translation problems can benefit us from at least two aspects:
	
	First, comparing to previous works only modeling the result consistency~\cite{DBLP:journals/tip/HeZKSC19,DBLP:conf/cvpr/ChoiCKH0C18,DBLP:conf/nips/ZhuZPDEWS17,DBLP:conf/icml/AlmahairiRSBC18,DBLP:conf/iccv/LinWCCL19,DBLP:conf/cvpr/GongLCG19,DBLP:conf/iccv/ZhuPIE17}, our method models two consistencies simultaneously,~\textit{i.e.}, result consistency and transition consistency. Explicitly modeling the transition between $x$ and $y$ and conducting learning under the transition consistency make the trained network model the inherent structure among closely related I2I translation tasks, which is proved in  comparison to previous representative works in {Fig.~\ref{fig:vip_face_editing}.(a)}. Since our method explicitly parameterizes the transition process and controls the consistency on both transition and results, our method outperforms previous I2I translation works focusing only on result consistency, which has been proved in our experiments.
	
	Second, to generalize our targeted result consistency and transition consistency to unseen transitions, we propose to explicitly model the unseen transitions $\tilde{t}$. In this way, the transition consistency defined on the observed transitions can be naturally extended to unseen transitions and its generated results. An example is presented in Fig.~\ref{fig:vip_face_editing}.(b).
	\begin{figure*}[t]
		\centering
		\includegraphics[width=16cm]{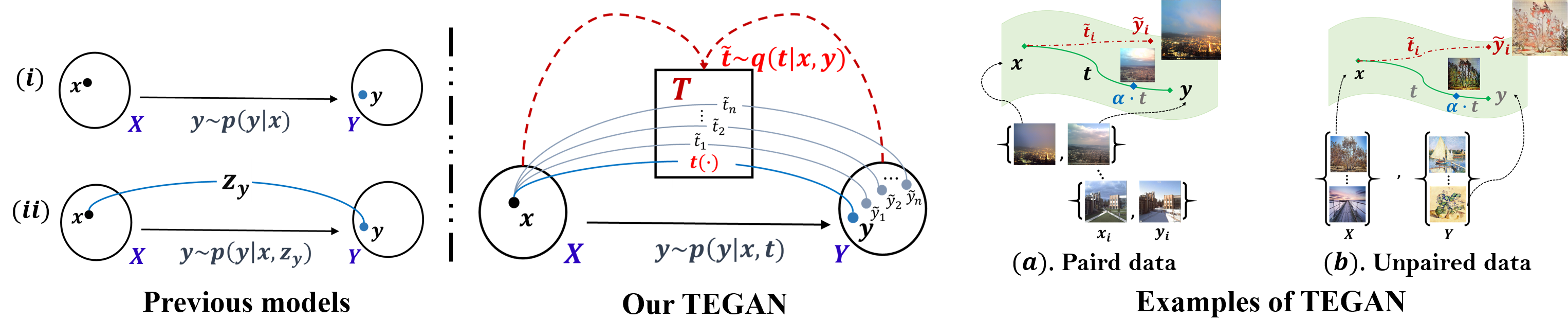}
		\vspace{-1mm}
		\caption{ \textbf{\textit{Left}:} Previous models focus on maintaining result consistency, with the data transformation implied within  {(\textit{i}) the training images or (\textit{ii}) attributes of the target output}, \textit{i.e.,} $z_{y}$. Failing to model unseen transitions, these models lack generalization ability to achieve high-quality translation results with unseen transitions, \textit{e.g.} Fig.~\ref{fig:vip_face_editing}.(a). \textbf{\textit{Middle}:} Our TEGAN consider both result consistency and transition consistency. We explicitly parameterize the transformation with transition {\small$t\triangleq t(x\cdot y)$ and model unseen transitions through encoding, \textit{i.e.,}{\small$\tilde{t}\sim q(t|x,y)$}. We further generalize the transition} consistency defined in observed transitions to unseen transitions with the designed constraints in Sec.~\ref{sec:limitation2}. Our TEGAN is a general generative I2I translation framework that can cover the existing models and also poses better generalization ability by regularize transition consistency for unseen transitions. \textbf{\textit{Right}:} Our TEGAN is general to be applied to various I2I translation tasks with either paired or unpaired data, \textit{e.g.} (a) supervised outdoor scene editing, (b) unsupervised style transfer.
		}\label{fig:motivation}
		\vspace{-4mm}
	\end{figure*}
	
	To overcome the limitation of simply regularizing results consistency on observed transitions %
	in existing methods, we explicitly model unseen transitions {\small$\tilde{t}$} with an transition encoding module~\cite{DBLP:journals/corr/KingmaW13}.
	which can flexibly manipulate $t$ to create reasonable new transitions for generation.  
	We define loss terms to enforce consistency between these explicitly mimicked triplet data.
	We then further generalized the overall consistency defined on the triplet data to the distribution-level via joint distribution matching~\cite{donahue2016adversarial}. We name our method Transition Encoding GAN (TEGAN). 
	
	Our contributions are summarized as follows: 
	\begin{itemize}
		\vspace{-2mm}
		\item We propose TEGAN to model the result consistency and transition consistency simultaneously. To the best of our knowledge, this is the first time to explicitly model transition consistency in I2I translation problems. Comparing to previous works focusing on result consistency only, TEGAN can work in higher granularity, model dynamic relation between attributes, and generate more reasonable results with better qualities.
		\vspace{-2mm}
		\item TEGAN presents a general framework for I2I translation. As discussed in {Sec.~\ref{sec:sec_4_discussion}}, existing GAN-based I2I translation works can be connected and interpreted as special cases of our TEGAN.
		\vspace{-2mm}
		\item We conduct extensive experiments on various I2I translation problems, including face editing, image inpainting, style transfer.
		The experimental results demonstrate the efficacy and generality of TEGAN.
	\end{itemize}
	
	\section{Related Works}
	
	\noindent \textbf{GANs based I2I translation.} I2I translation aims to transform an input image from one domain X to another domain Y with different attribute(s)~\cite{hertzmann2001image}. Conditional Generative Adversarial Networks (cGANs) have been widely adopted to tackle I2I translation tasks and  have achieved impressive results. Specifically, some methods simply leverage example images for training, and formulate the problem as {\small$G(x) =\hat{y}\approx{y}$} where $\hat{y}$ is the generated output and {\small$\hat{y}\sim p(y|x)$}. For example, Pix2Pix~\cite{DBLP:conf/cvpr/IsolaZZE17} utilized paired data, {\textit{i.e.,}$\{x,y\}$} for training; CycleGAN\cite{DBLP:conf/iccv/ZhuPIE17} taking advantages of cycle-consistency to train with unpaired data, {\textit{i.e.,} {\small$x\in X$} and {\small$y\in Y$}}. Other works adopt attribute annotations of the target domain, denoted as $z_{y}$, to make the generated results present desired property, with the problem formulated as {{\small$G(x,z_{y})=\hat{y}\approx{y}$}, where {\small$\hat{y}\sim p(y|x,z_{y})$}} \cite{DBLP:journals/tip/HeZKSC19,DBLP:conf/cvpr/ChoiCKH0C18,DBLP:conf/nips/ZhuZPDEWS17,DBLP:conf/icml/AlmahairiRSBC18,DBLP:conf/iccv/LinWCCL19,DBLP:conf/cvpr/GongLCG19}.
	
	These previous methods focus on maintaining consistency defined on results, namely result consistency, for I2I translation. Such a design would restrict their translation capacity in two aspects. First, these models count on labeled data for training, either labeled with pair-wise images~\cite{lee2020maskgan} or extra attribute annotations of each images~\cite{choi2020stargan}. In this way, they can be largely restricted by the data collection difficulty~\cite{deng2020disentangled}. Second, since the result consistency can be simply defined on the training examples, such consistency can not be generalized to unseen transformation, \textit{e.g.,} {\small$ \tilde{x}\rightarrow\tilde{y}$}, to achieve reasonable results in the test phase. We present further discussions on these previous methods in Sec.~\ref{sec:sec_4_discussion}.
	
	
	\section{Method}\label{sec:sec2_generative_transition_mechanism}
	
	In this section, we firstly reformulate I2I translation problems with transition consistency, then we illustrate how to generalize TEGAN to unseen transitions; finally, we explain how to train TEGAN by designed loss terms.  
	\subsection{Looking back at I2I formulation with transition consistency}\label{sec:reformulateI2I}
	Let {\small$x\in{X}$} and {\small$y\in{Y}$} be the images of source domain {\small$X$} and target domain {\small$Y$}, with the mapping between each data pair characterized with a transition variable {\small{$t\triangleq {t(x,y)}\in{T}$}}, \textit{i.e.,} {\small$x \overset{t}{\mapsto}y$}. After introducing the transition variable $t$, I2I translation problems aim to learn a mapping $G$ such that $x$ can be transformed into the target $\hat{y}$, conditioned on transition $t$, {\textit{i.e.,} {\small$G(x,t)={\hat{y}}$}}.
	
	After reformulating I2I translation problems, the two consistencies in I2I translation should be formulated as: \begin{enumerate*}\item[{\small {\small 1)}}] result consistency: consistency defined on the output images,~\textit{i.e.}, $G(x,t)={\hat{y}}\approx y$; \item[{\small 2)}] transition consistency: the {necessary condition} that $y$ shares a common transition $t$ as the target image {\small$\hat{y}=G(x)$}, ~\textit{i.e.,}: 
		\vspace{-2mm}
	\end{enumerate*}
	\begin{small}
		\begin{equation}\label{eq:discrete_t}
		{(x,y)\rightarrow \hat{t}_{y}, \quad (x,\hat{y})\rightarrow \hat{t}_{\hat{y}}, \quad \hat{t}_{y}\approx{\hat{t}_{\hat{y}}}={t}, 
		}\end{equation}
	\end{small}
	where $\hat{t}_{y}$ and $\hat{t}_{\hat{y}}$ denote {the transition of} $\{x,y\}$ and $\{x,\hat{y}\}$, respectively. Eq.~\ref{eq:discrete_t} explicitly formulates the transition consistency: given a source data $x$, if its translated result $\hat{y}$ is close to $y$, then the transition from $x$ to $\hat{y}$, denoted as $\hat{t}_{\hat{y}}$, should be close to the transition from $x$ to $y$, denoted as  $\hat{t}_{y}$.
	\subsection{{{Generalize transition consistency to unseen data}}}\label{sec:limitation2}
	\begin{figure}[t]
		\begin{center}
			\includegraphics[width=0.3\textwidth]{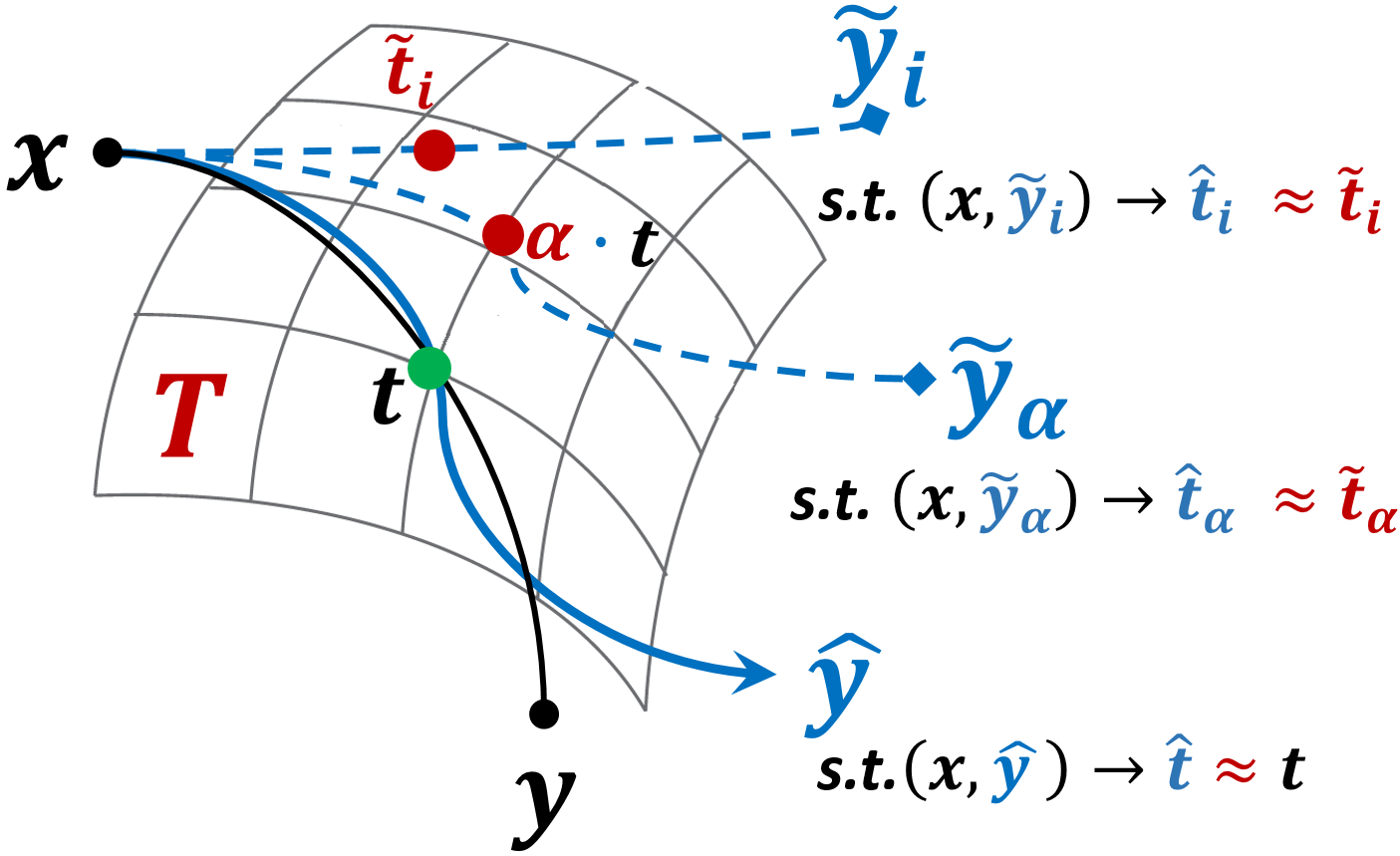}
		\end{center}
		\vspace{-4mm}
		\caption{\small {Our defined transition consistency for I2I translation.}
			We learn the distribution of $t$ with transition encoding, \textit{i.e.,} Eq.~\eqref{eq:learnable_t}, leading to a latent manifold denoted as $T$.
			We randomly sample {\small$\tilde{t}$} on the manifold, and define consistency between {\small$\tilde{t}$} and its corresponding generation {\small$\tilde{y}$}, with Eq.~\eqref{eq:sample_consistent_t}. We then generalize such transition consistency design with Eq.~\eqref{eq:consistent_t}. Our design facilitates {the definition and preservation} of translation consistency on unseen transitions. It also provides flexible and reasonable manipulation on $t$ with diverse operations. Interpolation adopted in previous works is one case of our design with simple linear operation. 
		}\label{fig:triplet_matching}
		\vspace{-4mm}
	\end{figure}
	Eq.~\eqref{eq:discrete_t} regularizes transition consistency for the observed transitions, \textit{i.e.,} {\small $x\overset{t}\rightarrow{y}$} in the training data. The generality for unseen transitions, denoted as $\tilde{t}$, depends on the quantity and diversity of training data. 
	
	To generalize the transition consistency to unseen transitions, we propose to 
	{explicitly model unseen transitions $\tilde{t}$} with transition encoding~\cite{DBLP:journals/corr/KingmaW13}, \textit{i.e.,} $x\times{y}\mapsto{t}$. Specifically, we learn a distribution of $t$ {with a ${q}$ function}, 
	The learned distribution characterizes a latent manifold $T$, where the observed transitions lied on. Random samples on this manifold are reasonable transitions $\tilde{t}_{i}$ that lead to new generation outputs, \textit{i.e.,}
	\vspace{-2mm}
	{\small$\tilde{y_{i}}=G(x,\tilde{t}_{i})$}. 
	\begin{equation}
	\begin{aligned}
	&q_{\phi}(t|x,y)\approx{p_{\theta}(t)}\approx{N(0,1)}, \\ 
	\quad \tilde{t}&\sim q_{\phi}(t|x,y), \qquad {t}'\sim{N(0,1)}. \label{eq:learnable_t}
	\end{aligned}
	\vspace{-1mm}
	\end{equation}
	
	In this way, the transition consistency on unseen transitions can then be explicitly regularized by  
	\begin{small}
		\begin{equation}
		\begin{aligned}
		{(x,\tilde{y})\rightarrow\hat{t}_{\tilde{y}},\ (x,{y}')\rightarrow\hat{t}_{{y'}},\  {\hat{t}_{\tilde{y}}\approx{\tilde{t}}},\ \hat{t}_{{y}'}\approx{t'}},\label{eq:sample_consistent_t}
		\end{aligned}
		\end{equation}        
	\end{small}
	where {\small${y}'=G(x,t')$}, {\small$\hat{t}_{\tilde{y}}$} and {\small$\hat{t}_{y'}$} denote the {transition} for {\small$\tilde{y}$} and {\small${y}'$}, respectively.
	
	Our design of transition encoding facilitates flexible manipulation on the transition, namely to obtain $\tilde{t}$ from $t$, with {both reasonable and diverse} operations. 
	First, our provided {\small$\tilde{t}$} are sampled from the learned distribution of $t$, making them imply inherent properties of the transformations.
	Second, random sampling on the manifold $T$, provides diverse operations \textit{w.r.t.} the given $t$. The interpolation adopted in previous works is one case of our design with simple linear operation, \textit{i.e.,} $\tilde{t}=\alpha\cdot{t}$, which is depicted in Fig.~\ref{fig:triplet_matching}. 
	\subsection{Consistency between unseen transition and its corresponding translation result}
	Although the consistency between sampled {\small$\tilde{t}$} and its translation result can be explicitly regularized through Eq.~\eqref{eq:sample_consistent_t}, such consistency can not be generalized to the transitions not sampled in the training phase. It is also infeasible to sample all transitions for explicit regularization. 
	Consequently, we further enforce a joint distribution matching on the triplet data, to generalize the consistency between $x$, $\tilde{t}$ and $\tilde{y}$ to distribution-level~\cite{DBLP:conf/cvpr/ZhengZZL0020}.
	\begin{small}
		\begin{equation}
		\begin{aligned}
		{{p_{\phi}(x, \tilde{t}, \tilde{y})}\approx {p_{\phi}(x, t', y')}\approx{p_{\phi}(x, \hat{t}, \hat{y})}  \approx{p_{\theta}(x,t,y)}}\label{eq:consistent_t}
		\end{aligned}
		\end{equation}        
	\end{small}
	With Eq.~\eqref{eq:consistent_t}, the transition consistency constraints defined in Eq.~\eqref{eq:sample_consistent_t} are further generalized to transitions that have not been explicitly generated (still unseen transitions) during training. This thus further benefits the model's generalization ability, facilitating it with even better translation performance in the test phase. We present Fig.~\ref{fig:triplet_matching} to illustrates our overall idea 
	to simultaneously tackle these issues. 
	
	\subsection{Our Model: Transition Encoding GAN}
	\begin{figure*}[t]
		\vspace{-7mm}
		\centering
		\includegraphics[width=15.5cm]{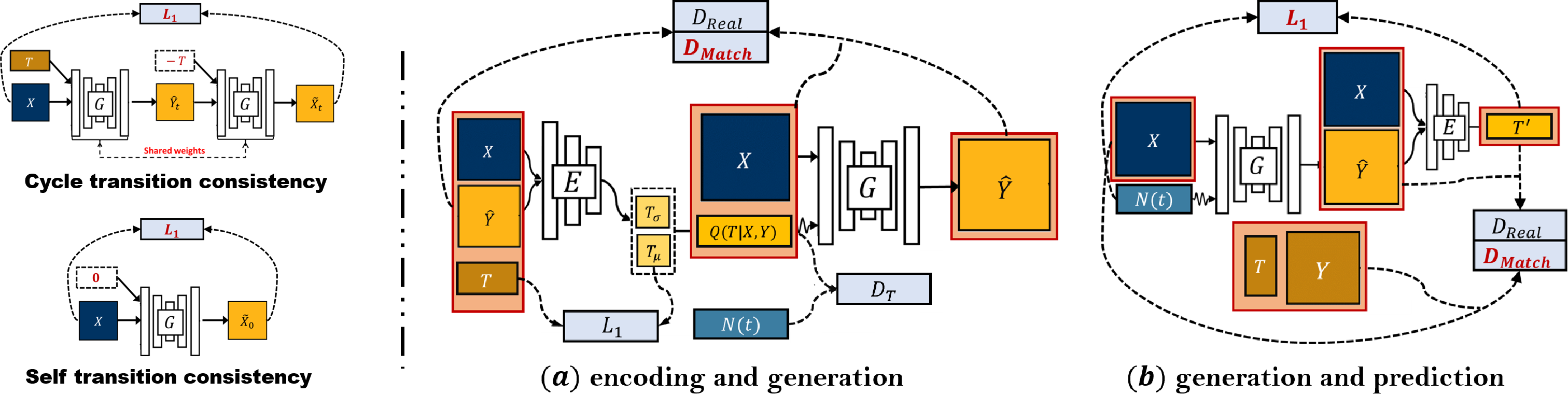}
		\caption{{The general network design of Transition Encoding GAN (TEGAN). \textbf{[Left]}. the design of cycle transition consistency and self-transition consistency. \textbf{[Right]}. the design of interactive generation consistency between encoding and generation.} }\label{fig:network}
		\vspace{-4mm}
	\end{figure*}
	Based on the above design, we present Transition Encoding GAN (TEGAN), which instantiate our idea with a stochastic encoder, \textit{i.e.,} {\small $q(t|x,y)\triangleq E(x,y)$}. 
	Note that, the general idea of our TEGAN can be applied to both the supervised setting and the unsupervised setting. 
	{Here, we discuss TEGAN in the unsupervised setting. Details of the supervised case of TEGAN are given in the supplementary.}
	
	As shown in Fig.~\ref{fig:motivation}, jointly modeling the result consistency and the transition consistency, our TEGAN is designed with two interactive generative processes.
	One is the vanilla translation process; the other is our proposed \textit{transition encoding process}. 
	The two generative process interact with each other through our design of generalized transition consistency on unseen transitions, \textit{i.e.,} Eq.\eqref{eq:sample_consistent_t} and Eq.\eqref{eq:consistent_t}.
	\vspace{-4mm}
	\subsubsection{Generation and reconstruction of image}\label{sec:model_generation_y}
	Our loss terms for {result consistency} in \textit{the vanilla translation process}, \textit{i.e.,} {\small$\hat{y}=G(x,t)$}, 
	includes: \vspace{1mm}\\
	\noindent\textbf{The adversarial loss for image generation.}
	\vspace{-2mm}
	\begin{small}
		\begin{equation}
		\begin{aligned}
		\min\limits_{G}\max\limits_{D_{\text{Real}}}\ &\mathcal{L}_{\text{Real\_img}} = \mathbb{E}_{y\sim{p(y)}}[\log D_{\text{Real}}(y)] \\
		+&\mathbb{E}_{(x,t)\sim{p(x,t)}}[\log (1- D_{\text{Real}}(G(x,t))] 
		\end{aligned}
		\end{equation}
	\end{small}
	where the generator $G$ aims to generate images that looks real, and the discriminator $D_{\text{Real}}$ aims to distinguish between the real images and the generated images. 
	
	\noindent\textbf{The losses for image reconstruction.} Based on the essence of our transition variable, we adopt the following loss terms for image reconstruction. \vspace{1mm}\\
	The {cycle-reconstruction loss is} 
	\begin{small}
		\begin{equation}
		\mathop{\min}\limits_{G}{\mathcal{L}_{\text{Recons\_img\_cyc}}} = \mathbb{E}_{(x,t)\sim{p(x,t)}}[\parallel G(G(x,t),-t)-x\parallel_{1}];
		\vspace{-2mm}
		\end{equation}
	\end{small}
	\noindent The {self-reconstruction loss} is  
	\begin{small}
		\begin{equation}
		\mathop{\min}\limits_{G}{\mathcal{L}_{\text{Recons\_img\_self}}}  = \mathbb{E}_{x\sim{p(x)}}[\parallel G(x,0)-x\parallel_{1}].
		\vspace{-2mm}
		\end{equation}
	\end{small}
	
	\vspace{-2mm}
	\subsubsection{Generation and reconstruction of transition}\label{sec:model_generation_t}
	Our losses for \textit{the transition encoding process} include
	
	\noindent\textbf{The adversarial loss for transition generation.}
	We adopt another discriminator on $t$, namely $D_{t}$, to learn a distribution for $t$ with our encoder, and consequently facilitate the generation of new transitions, \textit{i.e.,} \mbox{{\small$\tilde{t}\sim{q(t|x,y)}\triangleq E(x,y)$}}. 
	\begin{small}
		\begin{equation}
		\begin{aligned}
		\min\limits_{E}\max\limits_{D_{\text{t}}}\mathcal{L}_{\text{Real\_newtrans}} &= \mathbb{E}_{t\sim {p(t|x,y)}}[\log D_{\text{Lat}}(t)] \\
		\qquad & + \mathbb{E}_{t'\sim{N(t;0,1)}}[\log D_{\text{Lat}}(t')]\\
		\qquad &+ \mathbb{E}_{\tilde{t}\sim{q(t|x,y)}}[\log (1- D_{\text{t}}(\tilde{t})]
		\end{aligned}
		\end{equation}
	\end{small}
	\noindent\textbf{The loss for transition reconstruction.} The loss for {transition consistency} on observed transitions, \textit{i.e.,} Eq.~\eqref{eq:discrete_t}, is
	\begin{small}
		\begin{equation}
		\begin{aligned}
		\mathop{\min}\limits_{E}{\mathcal{L}_{\text{Recons\_trans}}}  &= \mathbb{E}_{(x,t,y)~\sim{p(x,t,y)}}[\parallel E(x,y)-t\parallel_{1}] \\
		\qquad &+\mathbb{E}_{(x,t,y)~\sim{p(x,t,y)}}[\parallel E(x,G(x,t))-t\parallel_{1}]\\
		\qquad &+\mathbb{E}_{x~\sim{p(x)}}[\parallel E(x,x)-0\parallel_{1}]
		\end{aligned}
		\end{equation}
	\end{small}
	\vspace{-4mm}
	\subsubsection{{Consistency losses for unseen translations}}\label{sec:model_interaction}
	The above two generation process cooperatively interact through our design of Eq.~\eqref{eq:sample_consistent_t} and Eq.~\eqref{eq:consistent_t},  \textit{i.e.,} enforce {\textbf{transition consistency}} on the generations facilitated with the unseen transitions. 
	The corresponding losses are
	\vspace{1mm}\\
	\noindent\textbf{The adversarial loss for new generation:}
	\begin{small}
		\begin{equation}
		\begin{aligned}
		\min\limits_{G}\max\limits_{D_{\text{{Real}}}}& \ \mathcal{L}_{\text{Real\_newimg}} = \mathbb{E}_{y\sim{p(y)}}[\log D_{\text{Real}}(y)] \\
		+&\mathbb{E}_{x\sim{p(x)},\tilde{t}\sim{q(t|x,y)}}[\log (1- D_{\text{Real}}(G(x,\tilde{t}))] \\
		+&\mathbb{E}_{x\sim{p(x)},\tilde{t'}\sim{N(0,1)}}[\log (1- D_{\text{Real}}(G(x,\tilde{t'}))] 
		\end{aligned}
		\end{equation}
	\end{small}
	\vspace{1mm}\\
	\noindent\textbf{The reconstruction of sampled transitions:}
	The reconstruction loss defined for transition consistency on the randomly sampled transitions, \textit{i.e.,} Eq.~\eqref{eq:sample_consistent_t} is
	\begin{small}
		\begin{equation}
		\begin{aligned}
		\mathop{\min}\limits_{E}{\mathcal{L}_{\text{Recons\_newtrans}}} & = \mathbb{E}_{\substack{x~\sim{p(x)}\\t'\sim{N(t;0,1)}}}{[\parallel E(x,G(x,t'))-t'\parallel_{1}]} \\
		\qquad &+\mathbb{E}_{\substack{x~\sim{p(x)}\\\tilde{t}~\sim{q(t|x,y)}}}{[\parallel E(x,G(x,\tilde{t}))-\tilde{t}\parallel_{1}]}
		\end{aligned}
		\end{equation}
	\end{small}
	\noindent\textbf{{The adversarial loss for triplet matching:}} We adopt a discriminator $D_{\text{Match}}$ that takes triplet inputs, \textit{i.e.,} $(x,t,y)$ to achieve our joint distribution matching design in Eq.~\eqref{eq:consistent_t}.
	\begin{small}
		\begin{equation}
		\begin{aligned}
		\min\limits_{G}\max\limits_{D_{\text{Match}}}\mathcal{L}_{\text{Match}} = &\;\;\mathbb{E}_{(x,t,y)\sim{p(x,t,y)}}[\log D_{\text{Match}}(x,t,y)] \\
		+\mathbb{E}_{(x,t)\sim{p(x,t)}}&[\log (1- D_{\text{Match}}(x, t, G(x,t))] \\
		+\mathbb{E}_{x\sim{p(x)},\tilde{t'}\sim{N(0,1)}}&[\log (1- D_{\text{Match}}(x, \tilde{t'}, G(x,\tilde{t'}))]\\
		+\mathbb{E}_{x\sim{p(x)},\tilde{t}\sim{q(t|x,y)}}&[\log (1- D_{\text{Match}}(G(x,\tilde{t},G(x,\tilde{t})))]
		\end{aligned}
		\end{equation}
	\end{small}
	{We further benefit triplet matching by incorporating wrong triplets $(x,t_{\times},y)$ and $(x,t,y_{\times})$, as in~\cite{DBLP:conf/iccv/LinWCCL19}.
		Our adversarial loss for triplet matching is then obtained as $\mathcal{L}^{D}_{\text{Match}}$ and $\mathcal{L}^{G}_{\text{Match}}$. (Pseudo-code given in the supplementary)}
	\vspace{1mm}\\
	\noindent \textbf{Full objective.} Finally, the full objective of our TEGAN is 
	\begin{small}
		\begin{equation}
		\begin{aligned}
		G',E'= \arg \min_{{G,E}}{\max_{D}}\;\;& 
		\mathcal{L}_{\text{Real\_img}} + \mathcal{L}_{\text{Real\_newimg}} + \mathcal{L}_{\text{Real\_newtrans}}\\
		+ &\lambda (\mathcal{L}^{D}_{\text{Match}} +\mathcal{L}^{G}_{\text{Match}})\\
		+ &\lambda_{\text{1}}(\mathcal{L}_{\text{Recons\_img\_cyc}}+\mathcal{L}_{\text{Recons\_img\_self}})\\ 
		+ &\lambda_{\text{2}}(\mathcal{L}_{\text{Recons\_trans}}+\mathcal{L}_{\text{Recons\_newtrans}})\label{eq:unsupervised_TEGAN}
		\end{aligned}
		\end{equation}
	\end{small}
	where $\lambda$, $\lambda_1$, $\lambda_2$ are hyper-parameters that control the relative importance of each term 
	respectively. 
	\subsection{Model training }
	{Fig.~\ref{fig:network}} presents the network design of our TEGAN. Specifically, modeling bi-directional triplet matching, our TEGAN is trained with two phases. In the (a) phase, we do encoding and generation, within which the generation and reconstruction on results and transitions are both constrained, \textit{i.e.,} the losses presented in Sec.~\ref{sec:model_generation_y} and Sec.~\ref{sec:model_generation_t}, respectively. In the (b), generation and encoding phase, we do reconstruction on the sampled transitions, i.e. Eq.~\eqref{eq:sample_consistent_t}. These two training phase are connected through the common discriminator $D_\text{Match}$ for the joint distribution matching of the triplet data, \textit{i.e.,} Eq.~\eqref{eq:consistent_t}. 
	\section{Discussion}~\label{sec:sec_4_discussion}
	Our proposed TEGAN provides a general generative framework for I2I translation. The aforementioned models can all be covered or explained with our TEGAN. We summarize their connections and comparisons in Tab.~\ref{tab:model_comparsion}.
	
	Generally, these methods simply model the result consistency on observed transitions, with the transition $t(x,y)$ implicitly conveyed within the training data. 
	For example,
	\textit{Pix2Pix}~\cite{DBLP:conf/cvpr/IsolaZZE17} use {\small$G(x,\cdot)={\hat{y}}\approx y$} as the constraint for result consistency, with $t$ implied within each data pair $x\rightarrow{y}$, while \textit{CycleGAN}~\cite{DBLP:conf/iccv/ZhuPIE17} tackles the problem with cycle-consistency~\cite{DBLP:conf/eccv/HuangLBK18}, \textit{i.e.,} {\small$G(G(x,t), -t)\approx{x}$}. 
	Other methods, \textit{e.g.} \textit{AttGAN}~\cite{DBLP:journals/tip/HeZKSC19} and \textit{StarGAN}~\cite{DBLP:conf/cvpr/ChoiCKH0C18}, regularize result consistency via attribute prediction on the translated image \textit{i.e.}{\small$\hat{y}\rightarrow{z_{y}}$}, with $t$ implied within $z_{y}$. 
	The self-reconstruction constraint in ~\cite{DBLP:conf/iccv/LinWCCL19} can be explained with transition as {\small$G(x, t = \mathbf{0})\approx{x}$}. 
	Without explicitly model $t(x,y)$, these methods can only regularize consistency on the observed transitions, limiting their generation capacity when translating with unseen transitions.
	
	\textit{RelGAN}~\cite{DBLP:conf/iccv/LinWCCL19}, \textit{DLOW}~\cite{DBLP:conf/cvpr/GongLCG19}, \textit{BicycleGAN}~\cite{DBLP:conf/nips/ZhuZPDEWS17} and \textit{AugCGAN}~\cite{DBLP:conf/icml/AlmahairiRSBC18} seeks to study consistency on unseen transitions. However, the generalization ability of their model could be inferior to 
	our TEGAN in mainly two aspects. \textit{RelGAN} and \textit{DLOW} introduce unseen transitions via synchronized interpolation on the observed $z_{y}$ and its corresponding image pair, \textit{i.e.,} {\small$G(x,\alpha\cdot t)\approx{(1-\alpha)\cdot x+\alpha\cdot y}$}. They can simply obtain {\small$\tilde{t}$} via interpolation, \textit{i.e.,} the simple linear case of {\small$\tilde{t}=\alpha\cdot{t}$} in the transition encoding of TEGAN (Fig.~\ref{fig:triplet_matching}), making their manipulation inflexible. In addition, the interpolated images may not be realistic itself, thus leading to unreasonable transitions {\small$\tilde{t}_{(*)}=t(x,\tilde{y}_{*})$} that fails to capture the intrinsic relations among the data, \textit{e.g.} relations between attribute annotations in face editing tasks~\cite{DBLP:conf/iccv/LinWCCL19}. This can disorder the transition consistency defined on unseen transitions. \textit{BicycleGAN}~\cite{DBLP:conf/nips/ZhuZPDEWS17} and \textit{AugCGAN}~\cite{DBLP:conf/icml/AlmahairiRSBC18} flexibly manipulate $t$ by encoding $z_{y}$, and enforce result consistency via attribute prediction. However, such regularization can only work on the explicitly sampled transitions. 
	
	Our TEGAN posses better generalization ability through jointly considering the result consistency and transition consistency for unseen transitions,
	and further generalizing both of them to distribution-level via joint distribution matching, \textit{i.e.} our Eq.~\eqref{eq:sample_consistent_t} and Eq.~\eqref{eq:consistent_t}, respectively.
	\begin{table}[t]
		\begin{center}
			\renewcommand{\arraystretch}{1.2}
			\caption{\small {Comparison and connections of different I2I translation methods with our proposed TEGAN.} 
			}
			\vspace{1mm}
			\label{tab:model_comparsion}
			\setlength{\tabcolsep}{1.18mm}{
				\scalebox{0.45}{
					\begin{tabular}{l|ccc|ccc|ccc}
						\toprule
						\multicolumn{1}{c|}{\multirow{2}{*}{\textbf{Methods}}} &
						\multicolumn{3}{c|}{\textbf{Transition}} &
						\multicolumn{3}{c|}{\textbf{Transition manipulation}} &
						\multicolumn{3}{c}{{\textbf{Consistency}}} \\ 
						\cline{2-10} 
						\multicolumn{1}{c|}{} & \multicolumn{1}{c|}{\begin{tabular}[c]{@{}c@{}}implied in\\{\small$x\rightarrow{y}$}\end{tabular}} &
						\multicolumn{1}{c|}{\begin{tabular}[c]{@{}c@{}}implied\\in {\small$z_{y}$}\end{tabular}} &
						\multicolumn{1}{c|}{\begin{tabular}[c]{@{}c@{}}\textbf{\textit{explicit}}\\{\small$t\triangleq t(x,y)$}\end{tabular}} &
						\multicolumn{1}{c|}{\begin{tabular}[c]{@{}c@{}}obtain {\small$\tilde{t}$}\\ from $t$ \end{tabular}} & 
						\multicolumn{1}{c|}{\begin{tabular}[c]{@{}c@{}}with \textbf{\textit{flexible}}\\operations\end{tabular}} &
						\multicolumn{1}{c|}{\begin{tabular}[c]{@{}c@{}}capture\\ \textit{\textbf{relations}}\\among\\{the data}\end{tabular}} &
						\multicolumn{1}{c|}{\begin{tabular}[c]{@{}c@{}}on\\observed\\transitions\end{tabular}} &
						\multicolumn{1}{c|}{\begin{tabular}[c]{@{}c@{}}on\\samples of\\ \textbf{\textit{unseen}}\\ \textbf{\textit{transitions}}\end{tabular}} &
						\multicolumn{1}{c}{\begin{tabular}[c]{@{}c@{}}generalized to\\ \textbf{distribution-level}\end{tabular}}\\ \hline
						\rowcolor{mygray1}
						Pix2Pix~\cite{DBLP:conf/cvpr/IsolaZZE17}     & \cmark & & & \xmark & \xmark &\xmark & \cmark&\xmark&\xmark\\
						\rowcolor{mygray1}
						CycleGAN~\cite{DBLP:conf/iccv/ZhuPIE17}      & \cmark & & & \xmark & \xmark& \xmark& \cmark&\xmark&\xmark\\
						\hline
						\rowcolor{mygray1}
						AttGAN~\cite{DBLP:journals/tip/HeZKSC19}     & & \cmark & & \xmark & \xmark& \xmark& \cmark&\xmark&\xmark \\
						\rowcolor{mygray1}
						StarGAN~\cite{DBLP:conf/cvpr/ChoiCKH0C18}    & & \cmark & & \xmark & \xmark& \xmark& \cmark&\xmark&\xmark \\
						\rowcolor{mygray2}
						RelGAN~\cite{DBLP:conf/iccv/LinWCCL19}       & &\cmark & &\cmark &\xmark& \xmark&  \cmark&\cmark&\xmark \\
						\rowcolor{mygray2}
						{DLOW~\cite{DBLP:conf/cvpr/GongLCG19}}       & &\cmark & &\cmark &\xmark& \xmark& \cmark&\cmark&\xmark\\
						\rowcolor{mygray3}
						BicycleGAN~\cite{DBLP:conf/nips/ZhuZPDEWS17} & & \cmark & & \cmark & \cmark & \cmark& \cmark&\cmark&\xmark\\
						\rowcolor{mygray3}
						AugCGAN~\cite{DBLP:conf/icml/AlmahairiRSBC18}& & \cmark & & \cmark & \cmark & \cmark& \cmark&\cmark&\xmark\\ 
						\hline
						TEGAN~{(ours)}                               & &&\cmark & \cmark & \cmark& \cmark&\cmark& \cmark&\cmark\\
						\bottomrule
			\end{tabular}}}
		\end{center}
		\vspace{-9mm}
	\end{table}
	\vspace{-3mm}
	\section{Experiments}\label{sec:sec_5_experiments}
	We test the performance of TEGAN on four I2I translation tasks, including face editing, outdoor scene editing, multi-domain style transfer, and image inpainting. 
	In these tasks, our transition posses different semantic definition, according to the desired attribute change in each of them.
	Our translation results are expected to be both realistic and transition consistent, \textit{i.e.,} presenting the property defined by $t$.
	\subsection{Face editing}~\label{sec:exp_face_editing}
	We first test TEGAN on face editing, where the transition $t$ specifies the change of facial attributes, \textit{i.e.,} {\small $x\overset{t = a_y-a_x}{\longrightarrow}{y}$}, where $a_y$ and $a_x$ denote the attribute annotation of face images $y$ and $x$, respectively. 
	\begin{figure}[t]
		\centering
		\includegraphics[width=7.3cm]{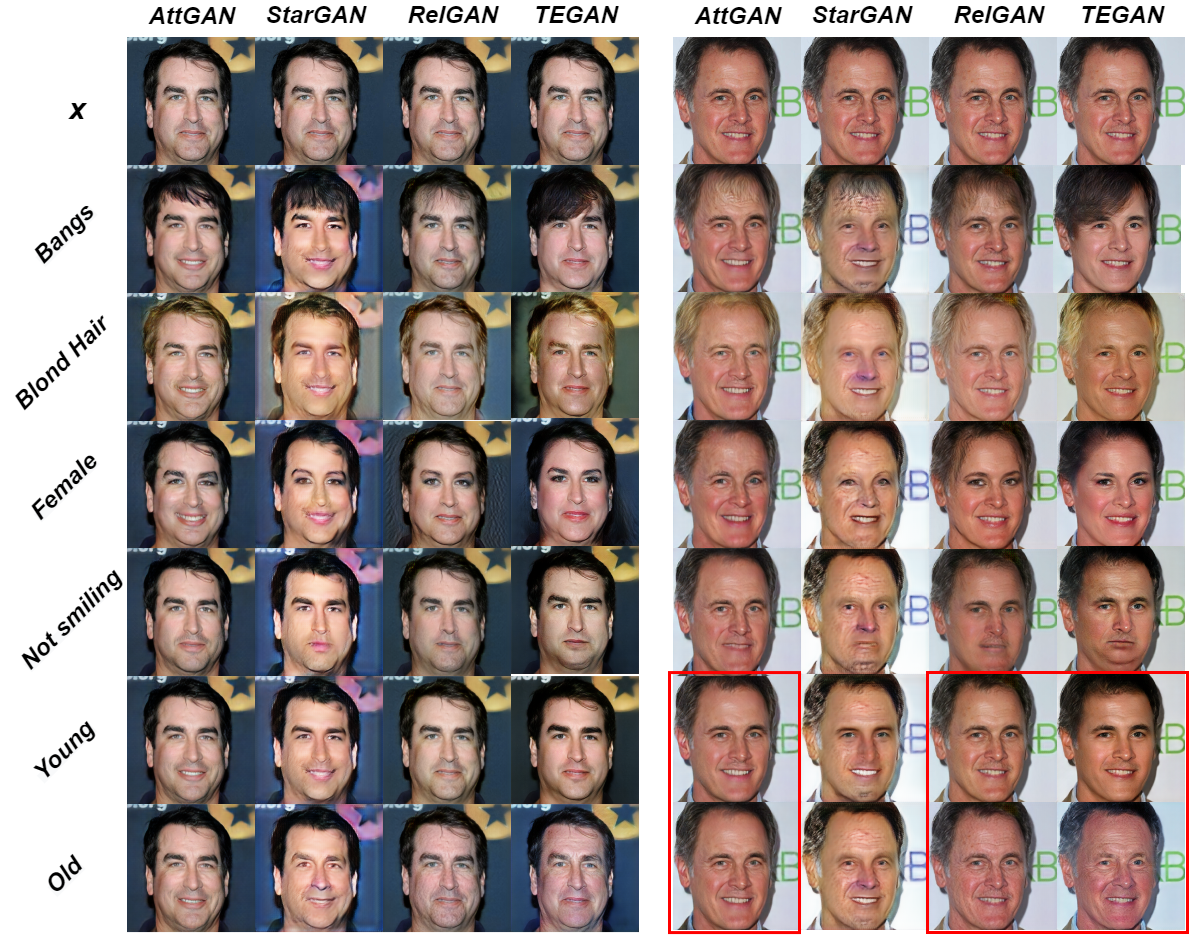} 
		\caption{Comparisons of TEGAN with AttGAN, StarGAN and RelGAN on binary facial attribute editing on CelebA-HQ dataset. More semantically meaningful results are marked with red rectangles. Zoom in for better resolution.}\label{fig:celeba_binary_editing} 
	\end{figure}
	\begin{figure}[t]
		\vspace{-0.3cm}
		\centering
		\includegraphics[width=7.3cm]{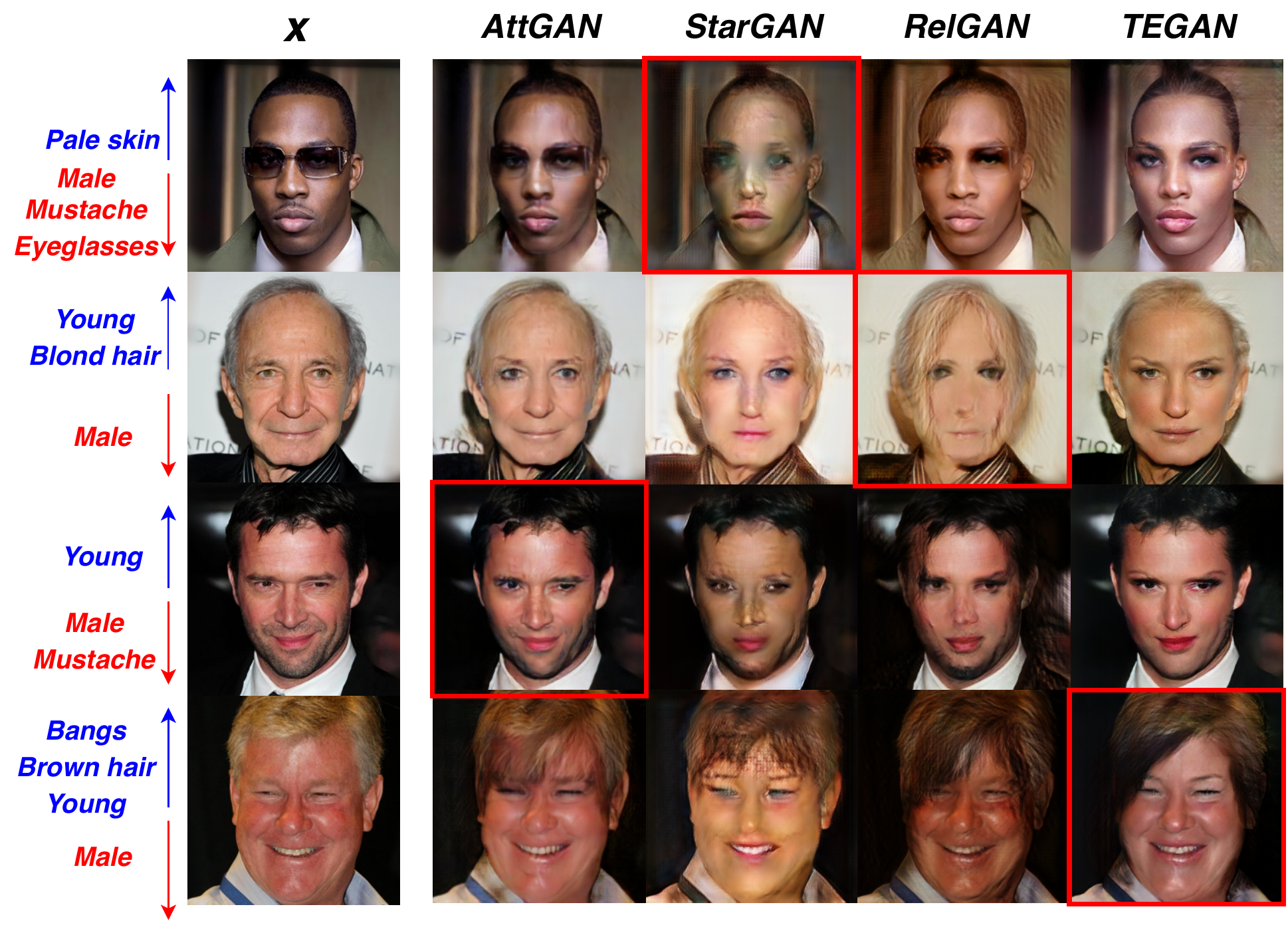}
		\caption{Comparisons results on \textit{multiple} facial attribute editing on CelebA-HQ dataset. More obvious comparisons are marked with red rectangles. Zoom in for better resolution.} \label{fig:celeba_multi_editing}
		\vspace{-6mm}
	\end{figure}
	
	\noindent \textbf{CelebA-HQ dataset:} 
	The CelebA-HQ is a High-Quality version of the CelebFaces Attributes (CelebA) dataset\footnote{https://github.com/switchablenorms/CelebAMask-HQ}. The dataset consists of 30,000 high-quality face images in CelebA at 1024 $\times$ 1024 resolution, each annotated with 40 binary attributes concerning hair colors, gender, age, etc.
	
	\noindent \textbf{Setting:} 
	We resize and center-crop these images to 256 $\times$ 256, and adopt 10 easily identifiable attributes, including hair color (black, blond, brown),  dress-up (eyeglasses, bangs, mustache, pale skin, smiling), gender(male/female) and age(young/old), for experiment. Among all these images, we adopt a 90/10 split for training and testing~\cite{DBLP:conf/iccv/LinWCCL19}. We compare with three state-of-the-art face editing methods: AttGAN~\cite{DBLP:journals/tip/HeZKSC19}, StarGAN~\cite{DBLP:conf/cvpr/ChoiCKH0C18} and RelGAN~\cite{DBLP:conf/iccv/LinWCCL19}.
	

	\noindent \textbf{Single attribute/seen transition:} 
	Fig.~\ref{fig:celeba_binary_editing} presents example comparison results of TEGAN and other baseline methods regarding single facial attribute editing. Our results are more semantically meaningful than AttGAN and RelGAN, while the results of StarGAN are the worst with notable artifacts in most images. In the first set(the first four columns) of comparison, all the results of baseline methods trigger the smiling attribute (\textit{Mouth} changes to be open), when \textit{smiling} is not the desired changed attribute. This is not semantically consistent with our desired changes. In the second set of results, our results present more evident visual effect of attributes compared with RelGAN and AttGAN. This phenomenon is most evident in the comparison between the editing results of \textit{young} and \textit{old} (marked with red rectangle in Fig.~\ref{fig:celeba_binary_editing}), where our TEGAN presents obviously meaningful semantics, especially the color of hairs. Overall, the results of our TEGAN is more realistic and semantically meaningful. We present more results in the supplementary, as well as the comparisons on binary attribute interpolations.
	
	\noindent \textbf{{{Multiple attributes/unseen transition:}}} 
	We further conduct experiments on multiple attributes editing task, where modeling unseen transitions are more important. 
	
	\noindent \textbf{Qualitative results:}
	As shown in Fig.~\ref{fig:celeba_multi_editing},  our TEGAN generates more realistic and semantically meaningful results among the four methods. To be specific, the results of StarGAN present clear deficiencies, especially the one marked with red rectangle in the first row. It is due to the mismatch between StarGAN's domain classifier and attribute annotations. RelGAN obtains more blurry results in this setting since it simply models the unseen transitions with linear operations, \textit{i.e.,} {\small$\tilde{t}=\alpha\cdot{t}$}, which can not cover other flexible combinations of attributes that do not present in the training set. 
	Although AttGAN achieves reasonable results, because these attributes are independently modeled, its generated results can not {present meaningful fusion of changes according to the inherent relation between the attributes, \textit{e.g.}, mustache still exists when we change the gender from male to female.(See the second image in the third row)} In contrast, results of TEGAN are more realistic and highly semantically meaningful, because of the ability of modeling unseen transitions in non-linear space. For example, TEGAN changes a \textit{male}'s \textit{bangs} when editing him to \textit{female} in the last image of the last row.
	
	\begin{table}[tbp]
		\vspace{-2mm}
		\begin{center}
			\renewcommand{\arraystretch}{1}
			\caption{\small	Fr\'echet Inception
				Distance (FID $\downarrow$) between input images and transformed images by three representative previous methods and TEGAN. Lower value means better. 
			}           \label{tab:multi_attrs_fid}
			\vspace{2mm}
			\setlength{\tabcolsep}{1mm}{
				\scalebox{0.98}{
					\begin{tabular}{c|cccc}
						\hline 
						\toprule
						& StarGAN & RelGAN &AttGAN &TEGAN  \\ \hline
						FID & 125.94 & 71.78  & 72.22 & \textbf{66.93} \\ 
						\bottomrule
					\end{tabular}
			}} \vspace{-9mm}
		\end{center}
	\end{table}
	
	\noindent \textbf{Quantitative results:}
	We qualitatively evaluate and compare the generation results. Table \ref{tab:multi_attrs_fid} presents the comparison regarding Fr\'echet Inception Distance {(FID)}~\cite{heusel2017gans} (lower is better) between the input image $x$ and the translated image $G(x,t)$. We can see TEGAN achieves the best results among all the methods, which indicates the translated images with high quality. Please refer to the supplementary for more results on multiple attributes editing and translated images by sampling unseen transitions.

	\subsection{Outdoor scenes editing}
	\label{sec:exp_night2day}
	\begin{figure}[t]
		\centering
		\includegraphics[width=7.3cm]{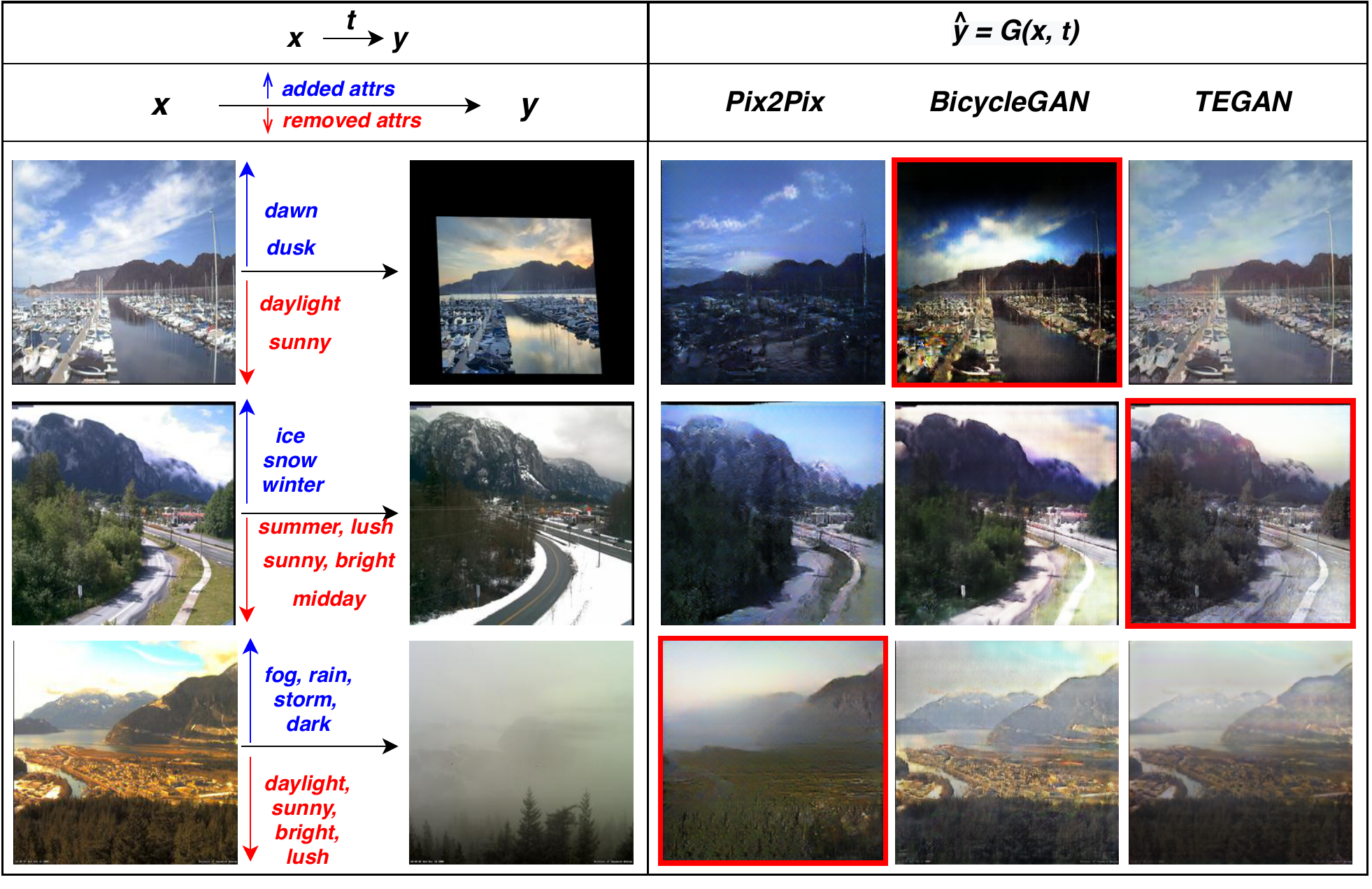}
		\caption{Comparisons of TEGAN with {Pix2Pix~\cite{DBLP:conf/cvpr/IsolaZZE17} and BicycleGAN~\cite{DBLP:conf/nips/ZhuZPDEWS17}} on supervised I2I translation on the Transient Attribute Dataset. More obvious results are marked with red rectangles. Zoom in for better resolution. }
		\label{fig:night2day_a2b}
		\vspace{-3mm}
	\end{figure} 
	We test the performance of TEGAN on supervised attribute transfer on the Transient Attribute Dataset~\cite{laffont2014transient}.\\
	\noindent \textbf{Transient Attribute Dataset:} It is a dataset containing scene images from 101 outdoor webcams. The images are captured over long time spans which exhibit drastic changes in appearance. The dataset contains 8571 images in total, each annotated with 40 transient attributes, such as \textit{sunny}, \textit{bright} and \textit{ice}. \\
	\noindent \textbf{Setting:} We select $20$ visually evident attributes to define the transition and randomly select one image taken from the same camera for each single image to construct $8571\times1$ paired images for experiment. $8076$ of these images are randomly selected for training, and the rest data are for testing. We adopt Pix2Pix and BicycleGAN\footnote{https://github.com/junyanz/BicycleGAN}, two state-of-the-art supervised I2I translation methods, for comparison.\\
	\noindent \textbf{Qualitative evaluation:} We present the qualitative comparison results in Fig.~\ref{fig:night2day_a2b}. Specifically, the results of baseline methods are less realistic than our TEGAN. For example, in the image marked with red rectangle in the last row, Pix2Pix fails to preserve the semantic properties declared in the ground truth images, \textit{i.e.} the river and houses are missing. In terms of the results of BicycleGAN in the first row, 
	BicycleGAN is influenced by deformation in the target image ($y$), with the details around the border of the image missing. In terms of semantic meanings presenting in the translated images, our results are more reasonable and semantically consistent with the declared transitions. In the second row of the figure, we expect the translated images would be a scene in winter with ice and snow. However, the translated images by Pix2Pix and RelGAN present more green color, which is not semantically consistent with the winter attribute. Overall, the results of our TEGAN present more realism and consistently semantic meanings. More results are given in the supplementary.\\
	\noindent \textbf{Quantitative evaluation:} We further compute the SSIM~\cite{bulat2018super} and PSNR~\cite{Zhang_2018_ECCV} scores of the generated images by each method for quantitative evaluation. The evaluated image pairs are ({\small {\small 1)}} input $x$ and $\hat{x}=G(x,0)$ (the reconstruction of $x$ with the $0$ transition); ({\small 2)} input $x$ and $\hat{y}=G(x,t)$ (the translated image of $x$ in domain $Y$).
	From Table. \ref{tab:ssim_night2day}, it is clear that our TEGAN achieves the highest score in all settings \textit{w.r.t.} the two metrics. 
	\begin{table}[t]
		\vspace{-4mm}
		\begin{center}
			\renewcommand{\arraystretch}{1}
			\caption{\small Comparison of the SSIM($\uparrow$) and PSNR($\uparrow$) of the translated images for the Transient Attribute Dataset. The best results are highlighted in bold. The larger the better.}
			\vspace{1mm}
			\label{tab:ssim_night2day}
			\setlength{\tabcolsep}{0.8mm}{
				\scalebox{0.85}{
					\begin{tabular}{c|c|ccc}
						\toprule
						{Metrics}      & {Image Pairs}             & {Pix2Pix}   & {BicycleGAN}   & {TEGAN\textit{(ours)}} \\ \hline
						\multirow{3}{*}{SSIM} & $x$ and $\hat{x}=G(x,0)$            &    -               &     0.52              &        \textbf{0.82}     \\
						& $x$ and $\hat{y}=G(x,t)$            &    0.25            &     0.41              &        \textbf{0.52}      \\
						\hline
						\multirow{3}{*}{PSNR} & $x$ and $\hat{x}=G(x,0)$            &    -               &     14.05             &        \textbf{22.52}     \\
						& $x$ and $\hat{y}=G(x,t)$            &    8.5             &     10.41             &        \textbf{12.36}     \\
						\bottomrule
			\end{tabular}}}
		\end{center}
	\end{table}

	\begin{figure}[t]
		\centering
		\includegraphics[width=7.3cm]{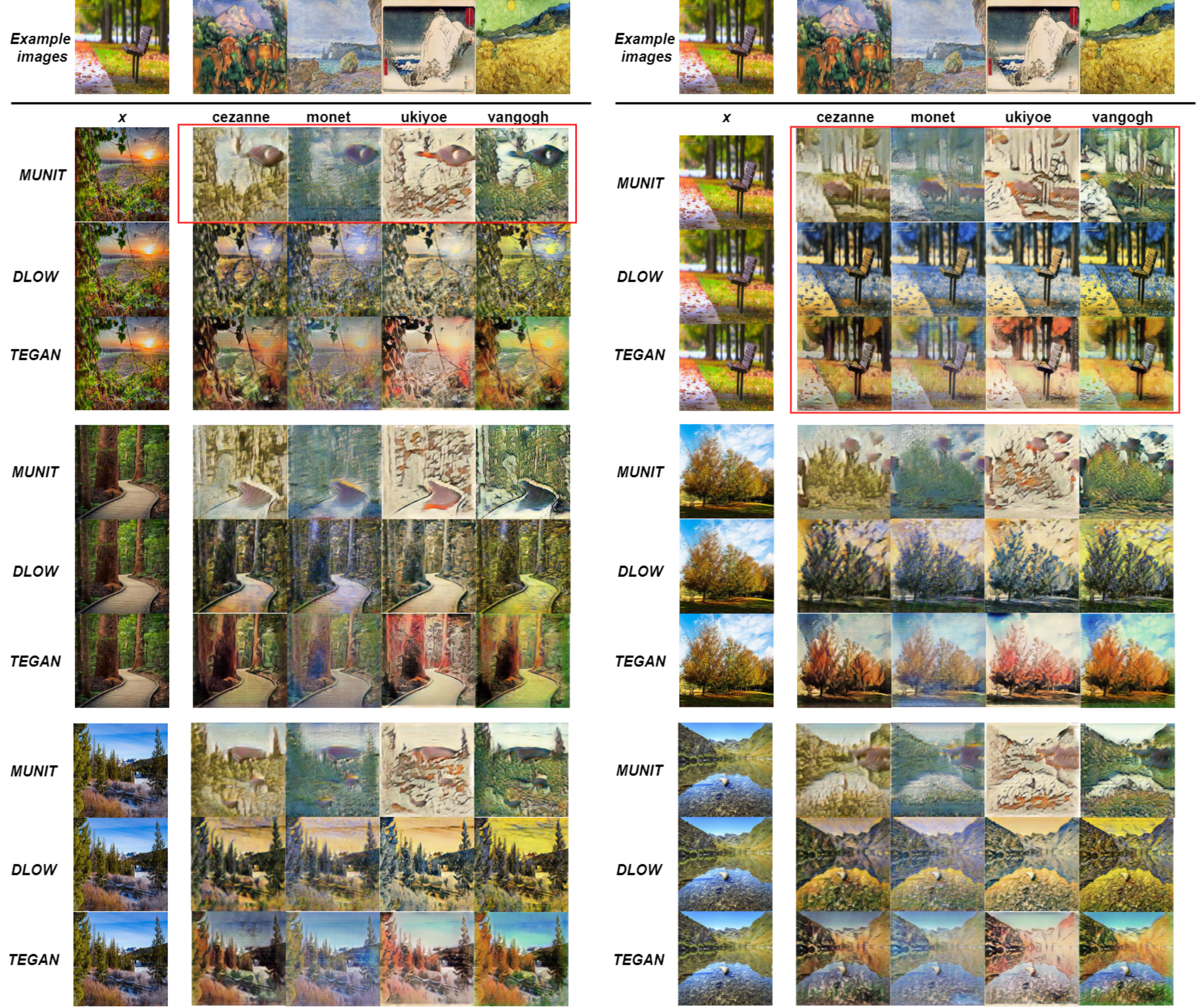}
		\caption{Comparisons of TEGAN with DLOW~\cite{DBLP:conf/cvpr/GongLCG19} on multi-domain style transfer. Images marked with red rectangles from DLOW present flaws as a fusion of different styles(purple color from monet). Zoom in for better resolution. } \label{fig:styletransfer_editing}
		\vspace{-5mm}
	\end{figure} 
	\subsection{Multi-domain style transfer}~\label{sec:exp_style_transfer} 
	Due to the flexibility of the semantics conveyed by transition, our TEGAN is handy to be generalized to conduct image-to-image translation with multiple target domains {\small $Y^{\{1,2,...n\}}$}, namely Multi-Domain I2I translation, with $t$ specifies the target domain index of each transformation, \textit{i.e.,} {\small $x\overset{t}{\longrightarrow}{y^{t}}$}. 
	Here, we adopt multi-domain style transfer as an example to testify TEGAN's ability in multi-domain I2I translation tasks. 
	The translation results are expected to be realistic and present obviously distinguishable visual characteristics for each target domain.
	
	\noindent \textbf{Photo2Art dataset:} There are 4 commonly adopted photo$\rightarrow$artistic painting datasets for style transfer task, \textit{i.e.,} \textit{photo$\rightarrow$cazanne}, \textit{photo$\rightarrow$monet}, \textit{photo$\rightarrow$ukiyoe} and \textit{photo$\rightarrow$vangogh}. For each photo example $x$, we randomly sample one art painting as $y$ and consequently constitute one triplet data sample, where $t$ specifies the style index of the target output. With all the photo images in these datasets, we finally construct a \textit{Photo2Art dataset} containing a total number of $25148$ \textit{photo$\rightarrow$art} image pairs. 
	
	\noindent \textbf{Setting:} We randomly select $500$ samples among the whole dataset for testing, and use the rest of the data for training. We adopt DLOW~\cite{DBLP:conf/cvpr/GongLCG19} as the baseline method.
	
	\noindent \textbf{Qualitative evaluation:}  Fig.~\ref{fig:styletransfer_editing} presents example comparisons regarding style transfer to each single domain. The results of our TEGAN present distinguishable stylization for each input, with each generated image presents recognizable style of the target domain. The results of DLOW fail to present clear distinction between the stylizaion of different domains. For example, the images in the second row present similar colors, and the same effect is shown in the fourth row. Besides, compared with the example images in the first row, some of results from DLOW also present evident flaw as a fusion of different styles. As shown in the figure, the images marked with red rectangles present some purple color pixels which is the representative color of \textit{monet} style. The reason is that DLOW facilitates model generalization by providing interpolated intermediate transitions for simple mix-up~\cite{DBLP:conf/iclr/ZhangCDL18} of the existing training data, and constrain the transition consistency by preserving distance proportion, \textit{i.e.,} the domainness variable $z$. In this way, the mix-up design may disorder the model from generating images with clear target property, \textit{i.e.,} $z=1$. This shows the priority of our transition encoding.
	\subsection{Image inpainting}
	\label{sec:exp_inpainting}
	\begin{figure}[t]
		\centering
		\includegraphics[width=7.3cm]{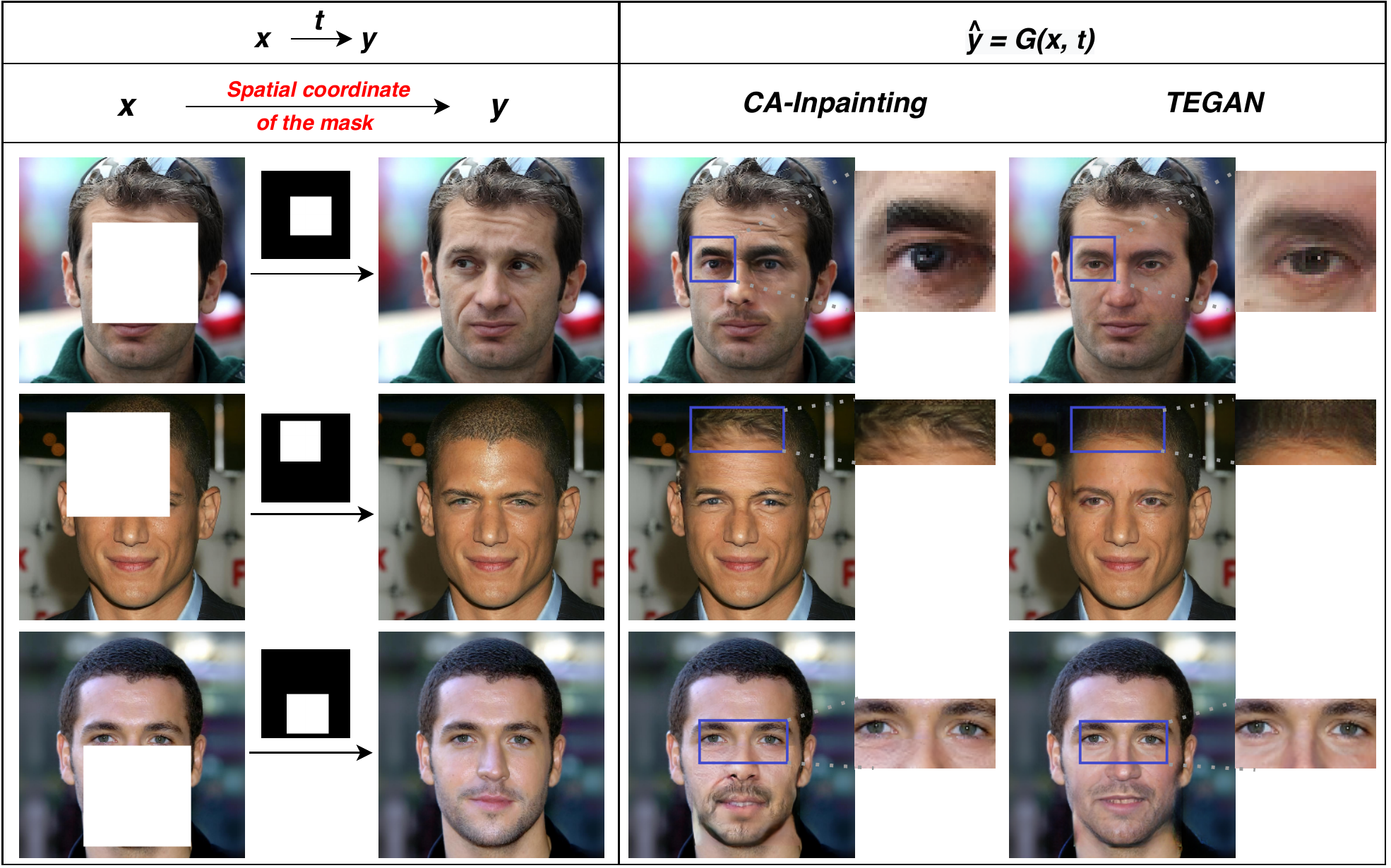}
		\caption{Comparisons of our TEGAN with CA-Inpainting for image inpainting on the CelebA-HQ dataset. The regions marked with blue rectangles are highlighted for detailed comparison. Best viewed with zoom-in.} \label{fig:inpainting_a2b}
	\end{figure}
	We additionally apply our TEGAN to image inpainting task, where $t$ conveys the change of relative position between the realistic image region and the missing region, \textit{i.e.,} clues of the masking regions, in each masked input. The translation results should be realistic and plausible to filling the missing regions. More importantly, less evident boundaries are expected in our TEGAN, due to the enforced consistency between the region of mask and region of the in-painted pixels in the generated output. 
	
	\noindent \textbf{CelebA-HQ dataset:} We adopt images in the aforementioned CelebA-HQ dataset to conduct image inpainting task. We resize the original $1024 \times 1024$  images to $256 \times 256 $ and erase an $128\times128$ squared region with an randomly sampled spatial location.  Thus we get the inpainting dataset for TEGAN, where $x$ is the marked incomplete image, $y$ is the originally complete image, and $t$ specifies the masked region in this data sample. 
	We randomly select $1,000$ samples among the whole dataset for testing; the rest of the data for training. 
	
	\noindent \textbf{Setting:} Following the state-of-the-art generative inpainting methods, we equip our TEGAN with an additional local critic, \textit{i.e.,} discriminator, to pay more concern on the quality of the inpainted regions. {To clarify the semantics of transition in inpainting task, we expand $t$ with one extra dimension to specify the translation direction in TEGAN, \textit{i.e.,} $+1$ indicates the positive direction to inpaint the masked images to complete ones, while $-1$ means the inverse direction, \textit{i.e.,} masking the complete image to the corresponding incomplete ones. }
	We adopt the model of generative image Inpainting with Contextual Attention\footnote{https://github.com/JiahuiYu/generative\_inpainting}~\cite{DBLP:conf/cvpr/Yu0YSLH18} (named as CA-Inpainting for simplicity), for a comparison.
	
	
	
	\noindent \textbf{Qualitative evaluation:} Example results of the inpainting task is presented in Fig.~\ref{fig:inpainting_a2b}. All these results are direct outputs of each model without any post-processing step. Considering the overall image quality, our TEGAN achieves comparable results with the CA-inpainting, the state-of-the-art method specialised in inpainting.
	However, the generated images of CA-inpainting suffers from sharp changes along the boundary regions of the mask. This is shown in the marked regions with blue rectangles in Fig.~\ref{fig:inpainting_a2b}. For example, in the first row of figure, CA-Inpainting just generate parts of the eyebrows. The similar phenomenons are also shown in the other example images. 
	Considering these boundary cases, the generated results of our TEGAN is much better, since its inpainted regions present a harmonious integration with the given context images. The reason behind our advantage is that our TEGAN poses better generalization ability to inpainting with unseen masking regions through our generative transition encoding design. 
	
	\section{Conclusion}
	This paper lodges a generative transition mechanism, TEGAN, to model the transition consistency among generated I2I translations.  Our TEGAN not only covers existing GAN-based I2I translation models, but also shows superior generalization ability benefited from our {consistent generative transition} design on unseen transitions.

	{\small
		\bibliographystyle{ieee_fullname}
		\bibliography{egbib}

\begin{thebibliography}{10}\itemsep=-1pt

\bibitem{DBLP:conf/icml/AlmahairiRSBC18}
Amjad Almahairi, Sai Rajeswar, Alessandro Sordoni, Philip Bachman, and Aaron~C.
  Courville.
\newblock Augmented cyclegan: Learning many-to-many mappings from unpaired
  data.
\newblock In Jennifer~G. Dy and Andreas Krause, editors, {\em Proceedings of
  the 35th International Conference on Machine Learning, {ICML} 2018,
  Stockholmsm{\"{a}}ssan, Stockholm, Sweden, July 10-15, 2018}, volume~80 of
  {\em Proceedings of Machine Learning Research}, pages 195--204. {PMLR}, 2018.

\bibitem{bulat2018super}
A. Bulat and G. Tzimiropoulos.
\newblock Super-fan: Integrated facial landmark localization and
  super-resolution of real-world low resolution faces in arbitrary poses with
  gans.
\newblock In {\em CVPR}, pages 109--117, 2018.

\bibitem{DBLP:conf/cvpr/ChoiCKH0C18}
Yunjey Choi, Min{-}Je Choi, Munyoung Kim, Jung{-}Woo Ha, Sunghun Kim, and
  Jaegul Choo.
\newblock Stargan: Unified generative adversarial networks for multi-domain
  image-to-image translation.
\newblock In {\em 2018 {IEEE} Conference on Computer Vision and Pattern
  Recognition, {CVPR} 2018, Salt Lake City, UT, USA, June 18-22, 2018}, pages
  8789--8797. {IEEE} Computer Society, 2018.

\bibitem{choi2020stargan}
Yunjey Choi, Youngjung Uh, Jaejun Yoo, and Jung-Woo Ha.
\newblock Stargan v2: Diverse image synthesis for multiple domains.
\newblock In {\em Proceedings of the IEEE/CVF Conference on Computer Vision and
  Pattern Recognition}, pages 8188--8197, 2020.

\bibitem{deng2020disentangled}
Yu Deng, Jiaolong Yang, Dong Chen, Fang Wen, and Xin Tong.
\newblock Disentangled and controllable face image generation via 3d
  imitative-contrastive learning.
\newblock In {\em Proceedings of the IEEE/CVF Conference on Computer Vision and
  Pattern Recognition}, pages 5154--5163, 2020.

\bibitem{donahue2016adversarial}
Jeff Donahue, Philipp Kr{\"a}henb{\"u}hl, and Trevor Darrell.
\newblock Adversarial feature learning.
\newblock {\em arXiv preprint arXiv:1605.09782}, 2016.

\bibitem{DBLP:conf/cvpr/GatysEB16}
Leon~A. Gatys, Alexander~S. Ecker, and Matthias Bethge.
\newblock Image style transfer using convolutional neural networks.
\newblock In {\em 2016 {IEEE} Conference on Computer Vision and Pattern
  Recognition, {CVPR} 2016, Las Vegas, NV, USA, June 27-30, 2016}, pages
  2414--2423. {IEEE} Computer Society, 2016.

\bibitem{DBLP:conf/cvpr/GongLCG19}
Rui Gong, Wen Li, Yuhua Chen, and Luc~Van Gool.
\newblock {DLOW:} domain flow for adaptation and generalization.
\newblock In {\em {IEEE} Conference on Computer Vision and Pattern Recognition,
  {CVPR} 2019, Long Beach, CA, USA, June 16-20, 2019}, pages 2477--2486.
  Computer Vision Foundation / {IEEE}, 2019.

\bibitem{goodfellow2014generative}
Ian Goodfellow, Jean Pouget-Abadie, Mehdi Mirza, Bing Xu, David Warde-Farley,
  Sherjil Ozair, Aaron Courville, and Yoshua Bengio.
\newblock Generative adversarial nets.
\newblock In {\em Advances in neural information processing systems}, 2014.

\bibitem{DBLP:journals/tip/HeZKSC19}
Zhenliang He, Wangmeng Zuo, Meina Kan, Shiguang Shan, and Xilin Chen.
\newblock Attgan: Facial attribute editing by only changing what you want.
\newblock {\em {IEEE} Trans. Image Process.}, 28(11):5464--5478, 2019.

\bibitem{hertzmann2001image}
Aaron Hertzmann, Charles~E Jacobs, Nuria Oliver, Brian Curless, and David~H
  Salesin.
\newblock Image analogies.
\newblock In {\em Proceedings of the 28th annual conference on Computer
  graphics and interactive techniques}, pages 327--340, 2001.

\bibitem{heusel2017gans}
M. Heusel, H. Ramsauer, T. Unterthiner, B. Nessler, and S. Hochreiter.
\newblock Gans trained by a two time-scale update rule converge to a local nash
  equilibrium.
\newblock In {\em NIPS}, pages 6626--6637, 2017.

\bibitem{hu2020unsupervised}
Bingwen Hu, Zhedong Zheng, Ping Liu, Wankou Yang, and Mingwu Ren.
\newblock Unsupervised eyeglasses removal in the wild.
\newblock {\em IEEE Transactions on Cybernetics}, 2020.

\bibitem{DBLP:conf/eccv/HuangLBK18}
Xun Huang, Ming{-}Yu Liu, Serge~J. Belongie, and Jan Kautz.
\newblock Multimodal unsupervised image-to-image translation.
\newblock In Vittorio Ferrari, Martial Hebert, Cristian Sminchisescu, and Yair
  Weiss, editors, {\em Computer Vision - {ECCV} 2018 - 15th European
  Conference, Munich, Germany, September 8-14, 2018, Proceedings, Part {III}},
  volume 11207 of {\em Lecture Notes in Computer Science}, pages 179--196.
  Springer, 2018.

\bibitem{DBLP:conf/cvpr/IsolaZZE17}
Phillip Isola, Jun{-}Yan Zhu, Tinghui Zhou, and Alexei~A. Efros.
\newblock Image-to-image translation with conditional adversarial networks.
\newblock In {\em 2017 {IEEE} Conference on Computer Vision and Pattern
  Recognition, {CVPR} 2017, Honolulu, HI, USA, July 21-26, 2017}, pages
  5967--5976. {IEEE} Computer Society, 2017.

\bibitem{DBLP:journals/corr/KingmaW13}
Diederik~P. Kingma and Max Welling.
\newblock Auto-encoding variational bayes.
\newblock In Yoshua Bengio and Yann LeCun, editors, {\em 2nd International
  Conference on Learning Representations, {ICLR} 2014, Banff, AB, Canada, April
  14-16, 2014, Conference Track Proceedings}, 2014.

\bibitem{laffont2014transient}
Pierre-Yves Laffont, Zhile Ren, Xiaofeng Tao, Chao Qian, and James Hays.
\newblock Transient attributes for high-level understanding and editing of
  outdoor scenes.
\newblock {\em ACM Transactions on graphics (TOG)}, 33(4):1--11, 2014.

\bibitem{lee2020maskgan}
Cheng-Han Lee, Ziwei Liu, Lingyun Wu, and Ping Luo.
\newblock Maskgan: Towards diverse and interactive facial image manipulation.
\newblock In {\em Proceedings of the IEEE/CVF Conference on Computer Vision and
  Pattern Recognition}, pages 5549--5558, 2020.

\bibitem{DBLP:conf/iccv/LinWCCL19}
Yu{-}Jing Lin, Po{-}Wei Wu, Che{-}Han Chang, Edward~Y. Chang, and Shih{-}Wei
  Liao.
\newblock Relgan: Multi-domain image-to-image translation via relative
  attributes.
\newblock In {\em 2019 {IEEE/CVF} International Conference on Computer Vision,
  {ICCV} 2019, Seoul, Korea (South), October 27 - November 2, 2019}, pages
  5913--5921. {IEEE}, 2019.

\bibitem{DBLP:conf/aaai/PanL0YY20}
Pingbo Pan, Ping Liu, Yan Yan, Tianbao Yang, and Yi Yang.
\newblock Adversarial localized energy network for structured prediction.
\newblock In {\em {AAAI} 2020}, pages 5347--5354. {AAAI} Press, 2020.

\bibitem{DBLP:conf/cvpr/PathakKDDE16}
Deepak Pathak, Philipp Kr{\"{a}}henb{\"{u}}hl, Jeff Donahue, Trevor Darrell,
  and Alexei~A. Efros.
\newblock Context encoders: Feature learning by inpainting.
\newblock In {\em 2016 {IEEE} Conference on Computer Vision and Pattern
  Recognition, {CVPR} 2016, Las Vegas, NV, USA, June 27-30, 2016}, pages
  2536--2544. {IEEE} Computer Society, 2016.

\bibitem{DBLP:conf/iccv/YangDLYY19}
Zongxin Yang, Jian Dong, Ping Liu, Yi Yang, and Shuicheng Yan.
\newblock Very long natural scenery image prediction by outpainting.
\newblock In {\em ICCV 2019}, pages 10560--10569. {IEEE}, 2019.

\bibitem{DBLP:conf/cvpr/Yu0YSLH18}
Jiahui Yu, Zhe Lin, Jimei Yang, Xiaohui Shen, Xin Lu, and Thomas~S. Huang.
\newblock Generative image inpainting with contextual attention.
\newblock In {\em 2018 {IEEE} Conference on Computer Vision and Pattern
  Recognition, {CVPR} 2018, Salt Lake City, UT, USA, June 18-22, 2018}, pages
  5505--5514. {IEEE} Computer Society, 2018.

\bibitem{DBLP:conf/iclr/ZhangCDL18}
Hongyi Zhang, Moustapha Ciss{\'{e}}, Yann~N. Dauphin, and David Lopez{-}Paz.
\newblock mixup: Beyond empirical risk minimization.
\newblock In {\em 6th International Conference on Learning Representations,
  {ICLR} 2018, Vancouver, BC, Canada, April 30 - May 3, 2018, Conference Track
  Proceedings}. OpenReview.net, 2018.

\bibitem{Zhang_2018_ECCV}
Y. Zhang, K. Li, K. Li, L. Wang, B. Zhong, and Y. Fu.
\newblock Image super-resolution using very deep residual channel attention
  networks.
\newblock In {\em ECCV}, 2018.

\bibitem{DBLP:conf/cvpr/ZhengZZL0020}
Shuai Zheng, Zhenfeng Zhu, Xingxing Zhang, Zhizhe Liu, Jian Cheng, and Yao
  Zhao.
\newblock Distribution-induced bidirectional generative adversarial network for
  graph representation learning.
\newblock In {\em 2020 {IEEE/CVF} Conference on Computer Vision and Pattern
  Recognition, {CVPR} 2020, Seattle, WA, USA, June 13-19, 2020}, pages
  7222--7231. {IEEE}, 2020.

\bibitem{DBLP:conf/iccv/ZhuPIE17}
Jun{-}Yan Zhu, Taesung Park, Phillip Isola, and Alexei~A. Efros.
\newblock Unpaired image-to-image translation using cycle-consistent
  adversarial networks.
\newblock In {\em {IEEE} International Conference on Computer Vision, {ICCV}
  2017, Venice, Italy, October 22-29, 2017}, pages 2242--2251. {IEEE} Computer
  Society, 2017.

\bibitem{DBLP:conf/nips/ZhuZPDEWS17}
Jun{-}Yan Zhu, Richard Zhang, Deepak Pathak, Trevor Darrell, Alexei~A. Efros,
  Oliver Wang, and Eli Shechtman.
\newblock Toward multimodal image-to-image translation.
\newblock In Isabelle Guyon, Ulrike von Luxburg, Samy Bengio, Hanna~M. Wallach,
  Rob Fergus, S.~V.~N. Vishwanathan, and Roman Garnett, editors, {\em Advances
  in Neural Information Processing Systems 30: Annual Conference on Neural
  Information Processing Systems 2017, 4-9 December 2017, Long Beach, CA,
  {USA}}, pages 465--476, 2017.

\end{thebibliography}
	}
	
\end{document}